\documentclass{article} 
\usepackage{iclr2025_conference,times}
\usepackage{booktabs} 
\usepackage{multirow} 
\usepackage{graphicx} 
\usepackage{amsmath,amssymb,amsfonts}
\usepackage{algorithmic}
\usepackage{graphicx}
\usepackage{wrapfig}
\usepackage{textcomp}
\usepackage{xcolor}
\usepackage{subcaption}
\usepackage{booktabs}
\usepackage{multirow}
\usepackage{pifont}
\usepackage{colortbl}
\usepackage{subcaption}
\usepackage{caption} 
\usepackage{amsmath,amsfonts,bm}

\def\eqref#1{equation~\ref{#1}}

\def\1{\bm{1}}

\DeclareMathAlphabet{\mathsfit}{\encodingdefault}{\sfdefault}{m}{sl}
\SetMathAlphabet{\mathsfit}{bold}{\encodingdefault}{\sfdefault}{bx}{n}

\usepackage{hyperref}
\usepackage{url}

\title{Mitigating Cache Noise in Test-Time Adaptation for Large Vision-Language Models}

\author{
Haotian Zhai\textsuperscript{1,2,*}, 
Xinyu Chen\textsuperscript{2,*}, 
Can Zhang\textsuperscript{2,*}, 
Tianming Sha\textsuperscript{3}, 
Ruirui Li\textsuperscript{2,†} \\
\textsuperscript{1}University of Minnesota Twin Cities,\;
\textsuperscript{2}Beijing University of Chemical Technology,\;
\textsuperscript{3}Sun Yat-sen University \\
\texttt{zihan9198@gmail.com,\; centurychen1103@gmail.com,\; alexlessend@gmail.com,}\\
\texttt{shatm@mail2.sysu.edu.cn,\; liruirui@mail.buct.edu.cn}
}

\iclrfinalcopy 
\footnotetext{* These authors contributed equally to this work.}
\footnotetext{† Corresponding author.}

\begin{document}

\maketitle

\begin{abstract}
Test-time adaptation (TTA) of visual language models has recently attracted significant attention as a solution to the performance degradation caused by distribution shifts in downstream tasks. However, existing cache-based TTA methods have certain limitations. They mainly rely on the accuracy of cached feature labels, and the presence of noisy pseudo-labels can cause these features to deviate from their true distribution. This makes cache retrieval methods based on similarity matching highly sensitive to outliers or extreme samples. Moreover, current methods lack effective mechanisms to model class distributions, which limits their ability to fully exploit the potential of cached information. To address these challenges, we introduce a comprehensive and reliable caching mechanism and propose a novel zero-shot TTA method called ``Cache, Residual, Gaussian" (CRG). This method not only employs learnable residual parameters to better align positive and negative visual prototypes with text prototypes, thereby optimizing the quality of cached features, but also incorporates Gaussian Discriminant Analysis (GDA) to dynamically model intra-class feature distributions, further mitigating the impact of noisy features. Experimental results on 13 benchmarks demonstrate that CRG outperforms state-of-the-art TTA methods, showcasing exceptional robustness and adaptability.

\end{abstract}

\section{Introduction}

In recent years, vision-language models pre-trained on large-scale datasets, such as CLIP \cite{CLIP}, have demonstrated remarkable zero-shot capabilities in downstream tasks. However, the direct application of un-tuned CLIP in practical scenarios often yields suboptimal performance due to distributional shifts between training and downstream data. Moreover, obtaining even a small number of high-quality annotated samples can be impractical and costly in certain situations. Consequently, effectively adapting vision-language models in zero-shot settings has emerged as a significant research focus and technical challenge.
A series of Test-Time Adaptation (TTA) \cite{TPT, Feng2023DiverseDA, PromptAlign, TDA} methods have been introduced to dynamically address distribution shifts during the testing phase of vision-language models. Recently, a cache-based test-time adaptation method called TDA \cite{TDA} has been proposed. TDA maintains a lightweight cache during testing to store representative test samples, guiding the classification of subsequent samples.

\begin{figure}[!ht]
    \centering
    \begin{subfigure}[b]{0.48\textwidth} 
        \centering
        \includegraphics[width=\linewidth]{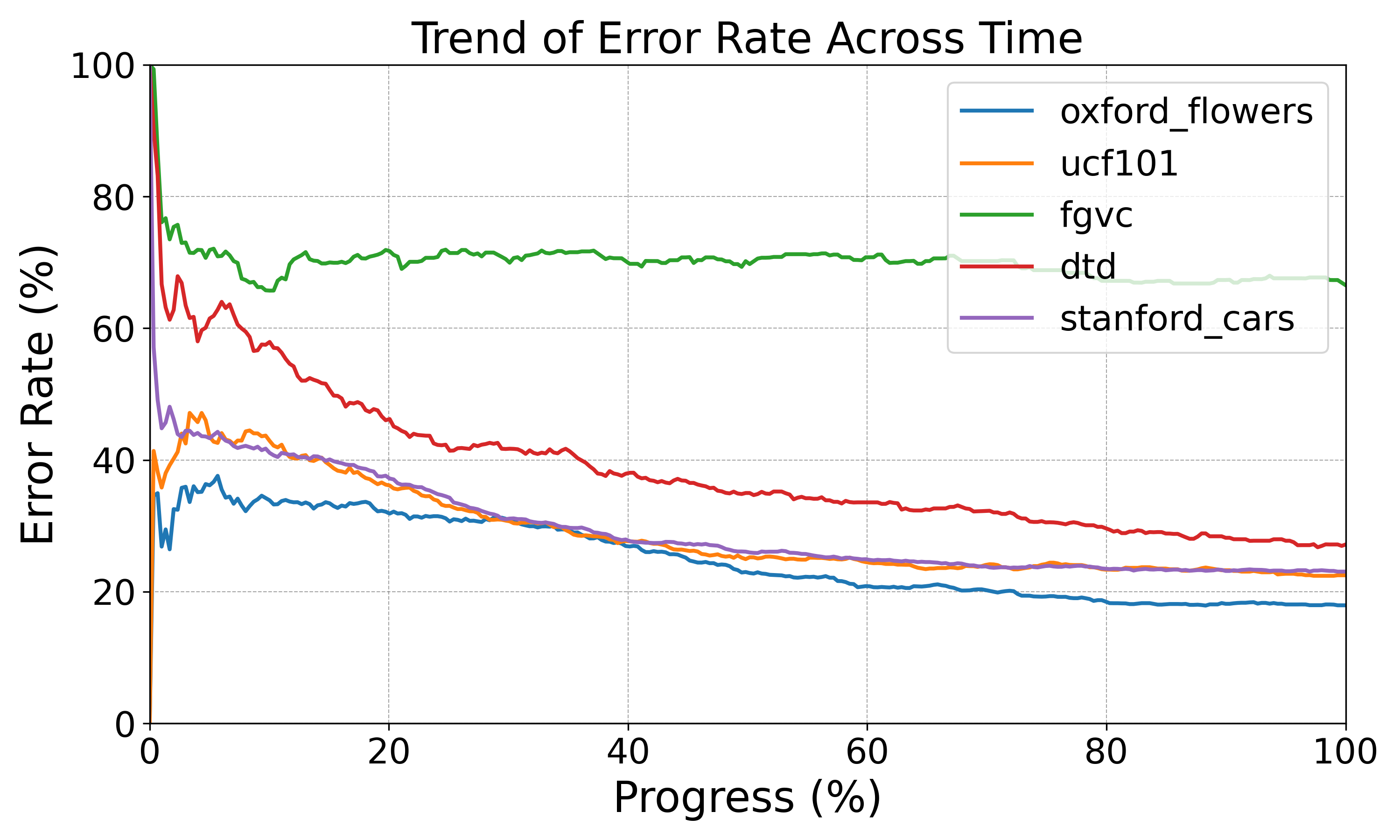}
        \caption{Error Rate Plot} 
        \label{fig:image11}
    \end{subfigure}
    \hfill 
    \begin{subfigure}[b]{0.48\textwidth} 
        \centering
        \includegraphics[width=\linewidth]{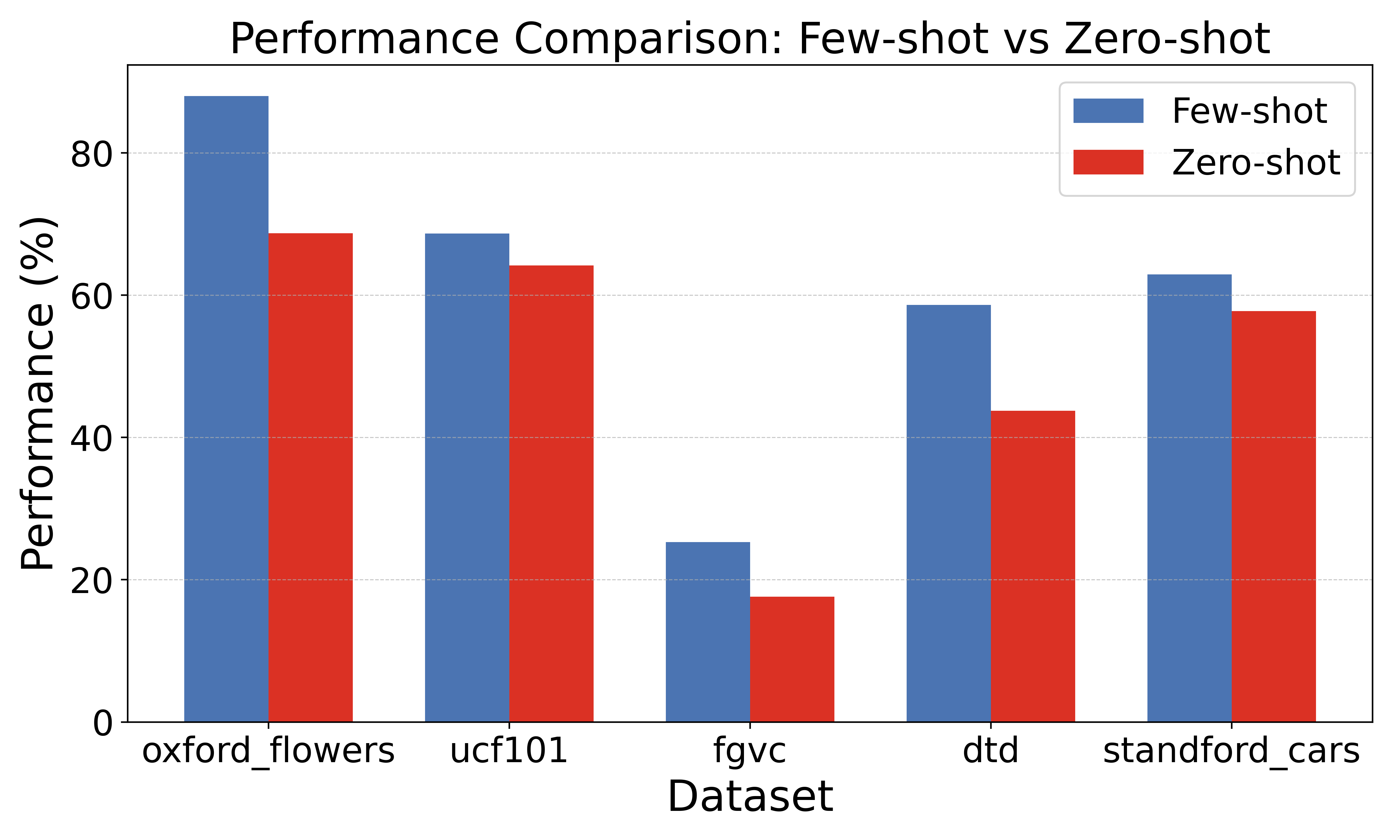}
        \caption{Few-shot vs Zero-shot Performance} 
        \label{fig:image22}
    \end{subfigure}
        \caption{Figure (a) shows the change in error rates for the positive cache during downstream testing with the TDA\cite{TDA} method, while Figure (b) compares the performance gap between similarity-based cache models in zero-shot(TDA\cite{TDA}) and few-shot (Tip-adapter\cite{TipAdapter}) settings, all using RN50 as the backbone.}
    \label{fig:comparison}
\end{figure}


We conducted an extensive review of the literature and further compared the performance of cache-based training-free methods through experiments. Tip-adapter\cite{TipAdapter} is a cache-based few-shot method, whereas TDA\cite{TDA} is a zero-shot method. Our observations revealed a significant performance gap between zero-shot and few-shot settings. Fig.~\ref{fig:image22} illustrates the performance of these methods under zero-shot and 8-shot conditions, demonstrating that the performance in the few-shot setting is significantly superior to that in the zero-shot setting. In these two configurations, the main difference lies in the content stored within the cache: the zero-shot method’s cache is populated using pseudo-labels. To further investigate, we analyzed the annotation accuracy of cached samples in the zero-shot setting, as shown in Fig.~\ref{fig:image11}. The error rate starts relatively high and decreases over the course of training but consistently remains above 20\%. We posit that these noisy labels contribute to an imbalance in the features stored in the cache and cause class prototypes to deviate from the true distribution. Unfortunately, previous research has not effectively addressed these noisy labels nor fully explored the distribution of features stored in the cache.


We introduce the Cache, Residual, Gaussian" (CRG) zero-shot test-time adaptation method to overcome these limitations, which features a comprehensive mechanism to enhance cache reliability. It includes two caches that separately store the positive and negative feature sets of image samples, along with an additional cache dedicated to storing text prototypes\cite{TPS}. The prototypes for the image features are calculated by averaging the features within the same class\cite{DPE}. To achieve thorough alignment between text and image prototypes, CRG employs learnable residual vectors \cite{DPE,TPS}, which are optimized by minimizing prediction entropy\cite{TPT} and maximizing inter-prototype distance. 

Despite achieving modality alignment, the cache may still harbor noisy features. To mitigate their effect, we model class distributions using the Gaussian Discriminant Analysis (GDA) framework \cite{bishop2006pattern, AHTB} and utilize Bayes' theorem to calculate posterior probabilities. This decision-making approach, grounded in distribution, effectively minimizes the impact of noisy samples. Additionally, to reduce the overconfidence in predictions induced by noisy features, we computed negative class prototypes, which serve as counter-references to further mitigate noise interference. Owing to the synergistic interaction of GDA and negative prototypes, CRG sustains heightened robustness and accuracy even in the presence of noise.


Our contributions can be summarized as follows:
\begin{itemize}
\item We have analyzed the factors limiting Test-Time Adaptation (TTA) performance and identified noisy labels as a key factor in the performance gap between zero-shot and few-shot settings.
\item We propose a novel TTA framework that effectively enhances robustness and generalization under noisy conditions by leveraging Gaussian Discriminant Analysis (GDA) and negative prototype learning.
\item During the TTA process, we simultaneously minimize prediction entropy and maximize inter-prototype distances, achieving effective prototype calibration based on learnable residuals, thereby improving overall adaptation performance.
\end{itemize}

\section{Related Works}
\subsection{Test-Time Adaptation for Vision-Language Models}
Due to distribution shifts between training and downstream data, many researchers have explored fine-tuning CLIP with a small amount of labeled data to mitigate these shifts and enhance adaptability. Existing CLIP fine-tuning methods can be divided into two main categories\cite{FewshotAO}: prompt-based methods (e.g., CoOp\cite{CoOp} and MaPLe\cite{MaPLe}) and adapter-based methods (e.g., Tip-Adapter\cite{TipAdapter} and CLIP-Adapter\cite{Gao2021CLIPAdapterBV}). Recently, test-time adaptation for large-scale vision-language models has attracted significant interest. TPT\cite{TPT} (Test-time Prompt Tuning) combines consistency regularization with prompt tuning to reinforce consistency among augmented views of test samples. DiffTPT\cite{Feng2023DiverseDA} builds on TPT, offering stronger augmentation capabilities at a higher computational cost. PromptAlign\cite{PromptAlign} employs the statistical features of a proxy dataset to align visual and textual representations. DPE\cite{DPE} introduces visual-textual prototype evolution to optimize modality alignment, while DMN\cite{DMN} utilizes a dynamic cache to store test-time prototype memories. TDA\cite{TDA} further enhances performance by incorporating a negative visual cache in addition to the positive visual cache.

\subsection{Negative Prototype Learning}
Prototype learning\cite{prototype} methods represent classes with prototypes and perform classification based on similarities to these prototypes. Many prototype learning approaches focus solely on visual modality tasks; for instance, DPNP\cite{DPNP} introduces a discriminative model built upon deep positive and negative prototypes. Decoupled Prototype\cite{wang2025decoupled} Learning applies positive and negative prototypes to online test-time adaptation, reducing performance degradation caused by noisy pseudo-labels. In visual-language models, SimNL\cite{simNL} enhances model adaptability in few-shot scenarios by constructing negative features in both the textual and visual modalities. In contrast, this paper introduces, for the first time, the concept of negative visual prototype learning for visual-language model test-time adaptation tasks. It aims to better align visual and textual prototypes in zero-shot scenarios by exploiting negative information at test time.

\section{Method}

Our work builds upon the CLIP visual-language model by incorporating the prototype residual vector learning technique from DPE\cite{DPE}. We innovatively propose a reliable Test-Time Adaptation (TTA) framework that includes caches, three sets of prototype residual learning, and decision-making based on Gaussian discriminant analysis.

\subsection{Preliminaries}

\textbf{CLIP\cite{CLIP}} (Contrastive Language-Image Pretraining) maps images and texts into a shared embedding space using an image encoder \( f \) and a text encoder \( g \). In the zero-shot classification scenario, for an image \( I \) and candidate classes \( \{c_1, c_2, \ldots, c_K\} \) with associated text prompts \( \{T_1, T_2, \ldots, T_K\} \), the probability of the image belonging to class \( c_k \) is computed as follows:

\[
p(c_k \mid I) = \frac{\exp\left( \tau \cdot \operatorname{sim}(f(I), g(T_k)) \right)}{\sum_{j=1}^K \exp\left( \tau \cdot \operatorname{sim}(f(I), g(T_j)) \right)}
\]

Here, the cosine similarity is defined as \(\operatorname{sim}(x, y) = \frac{x \cdot y}{\|x\| \|y\|}\), where: \( \tau \) is a temperature scaling parameter, \( K \) is the number of classes.

\subsection{Caches}


Inspired by the method of TDA\cite{TDA}, we adopt a priority queue strategy to store image features for each class. Specifically, for each test image \( I\), after making a prediction \(p(c_k \mid I)\), we assign it to the corresponding class queue based on its predicted pseudo-label \( \hat{y} \). Meanwhile, we compute the entropy value \( h_{I} \) of the test image and store the image feature together with its entropy as a pair \( (f(I), h_{I}) \) in the queue. For each class \( c_k \), the size of its priority queue is \( \mathcal{M} \), and each queue is initialized with a text feature \( g(T_k) \), the reasoning for which will be detailed in the Gaussian section.

When the queue is full, we compare the entropy value \( h_{I} \) of the new test image with the highest entropy value in the queue: if the new image has higher entropy, it will be discarded; if the entropy is lower, the image with the highest entropy in the queue will be replaced by the new image. The new image is inserted directly if the queue is not yet full.

We also construct a text-side cache \( T_{\text{cache}} \) to store class-specific text features derived from text descriptions. The cache has the shape \( T_{\text{cache}} \in \mathbb{R}^{K \times d} \), where
\[
T_{\text{cache}} = [t_1, t_2, \dots, t_K],
\]
and each \( t_k=g(T_k) \) represents the text feature for class \( c_k \). 

We update the text cache during inference to capture historical distribution information. We adopt DPE\cite{DPE}’s momentum update, using only high-confidence samples (entropy above threshold) to maintain stability.

\subsection{Residual}

Residual learning shows significant potential for enhancing model performance\cite{TPS, DPE, TaskRes}. More notably, we introduce learnable prototype residuals across the language, positive visual, and negative visual domains, enabling feature space calibration for each test sample during inference.

\textbf{Text Cache Residuals.}  
We use learnable residual parameters \( R_T \) to dynamically adjust the text cache during testing. Specifically, the text cache is updated as follows:

\begin{equation}
{T}_{\text{cache}} = \text{Normalize}({T}_{\text{cache}} + R_T)
\label{TCR}
\end{equation}

Here, \( R_T \in \mathbb{R}^{K \times d} \) is a learnable parameter initialized to zero.

\textbf{Positive Cache Residuals.}  
For each class \( c_k \), we compute the average of the features stored in the visual cache to obtain a cache prototype representing each class like DPE\cite{DPE}, denoted as \( V_{\text{cache}}^+ = [v_1^+, v_2^+, \dots, v_K ^+] \in \mathbb{R}^{K \times d} \). Subsequently, we use a set of visual residuals \( R_V^+ \) to further refine these prototypes. Specifically, the updated visual prototype is computed as:

\begin{equation}
    V_{\text{cache}}^+ = \text{Normalize}(V_{\text{cache}}^+ + R_V^+)
\end{equation}

where \( R_V^+ \in \mathbb{R}^{K \times d} \) is the learnable residual parameter, initialized to zero.

\textbf{Negative Cache Residuals.} Unlike TDA\cite{TDA}, which treats high-entropy samples as negative samples, our approach is based on the intuitive principle that if an image is classified as a dog, it cannot simultaneously be a cat—a concept inspired by SimNL\cite{simNL}. However, unlike both SimNL\cite{simNL} and TDA\cite{TDA}, we do not build a separate negative sample cache; instead, we directly extract negative prototypes from the existing positive visual prototypes, our method can be conceptually regarded as having a “virtual” negative cache.

To construct the negative cache prototypes \( V_{\text{cache}}^- = [v_1^-, v_2^-, \dots, v_n^-] \in \mathbb{R}^{K \times d} \), for each class \( c_k \), we compute the average of the prototypes from all other classes, effectively excluding class \( c_k \). Formally, the negative prototype is defined as:
\begin{equation}
{v}_k^{-} = \frac{1}{K-1} \sum_{\substack{j=1 \\ j \neq k}}^{K} v_j^{+}
\label{NCR}
\end{equation}

To further refine these negative prototypes, we introduce a set of learnable residual parameters \( R_V^{-} \in \mathbb{R}^{K \times d} \).  The updated negative prototypes are computed as:
\begin{equation}
V_{\text{cache}}^{-} = \text{Normalize}\left( V_{\text{cache}}^{-}  + R_V^{-} \right)
\end{equation}

where $V_{\text{cache}}^{-}$ is the set of the negtive protptypes and \( R_V^- \in \mathbb{R}^{K \times d} \) is the learnable residual parameter, initialized to zero.

\begin{figure*}[t]
    \centering
    \includegraphics[width=1\linewidth]{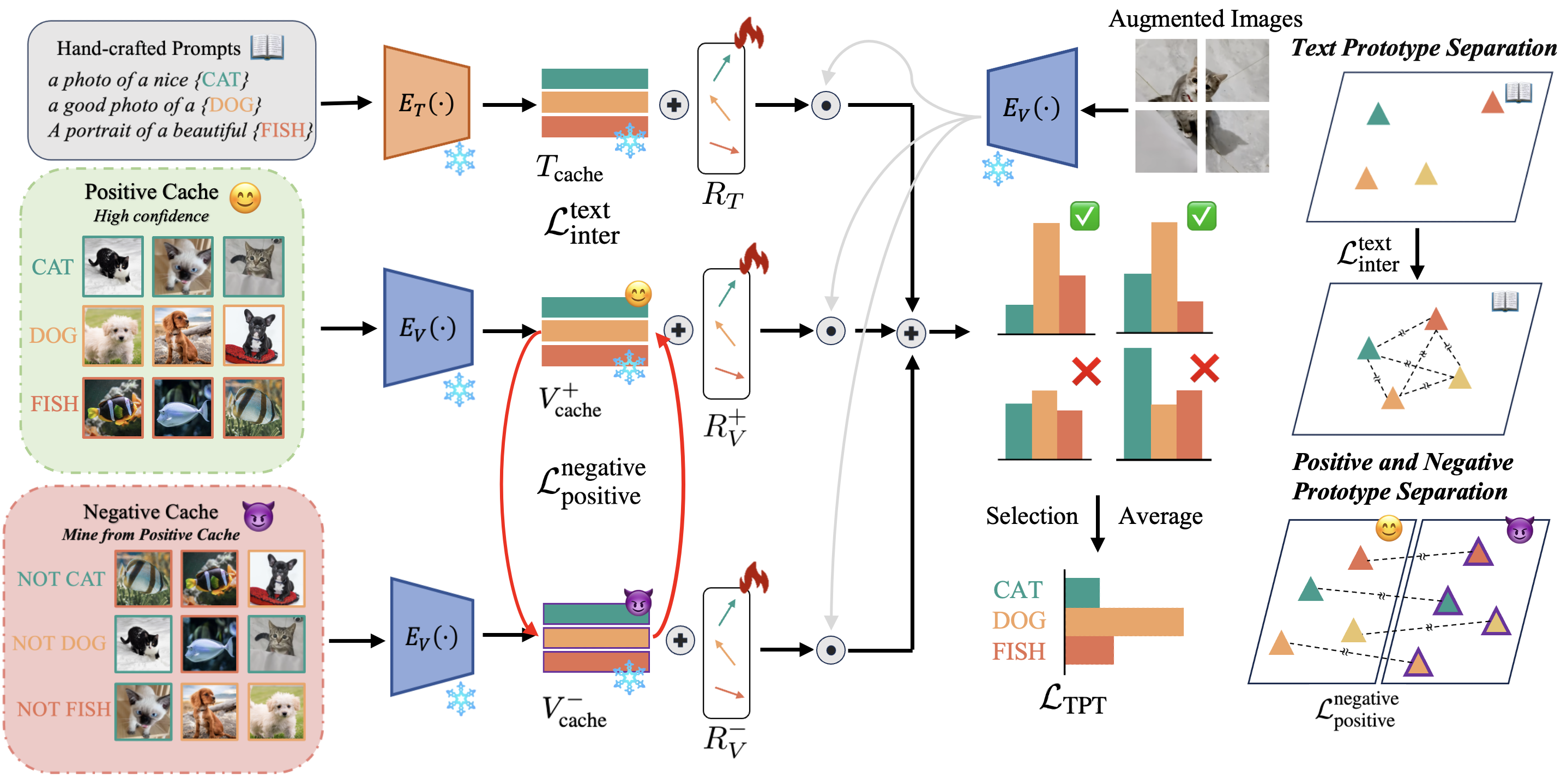}
    \caption{ \textbf{Overview of the CRG Method.} We introduce caches (both positive and negative) in the text and visual modalities and employ learnable residual vectors to flexibly calibrate multi-modal features. During inference, positive prototypes, negative prototypes, and text prototypes are used for similarity matching, while GDA models the feature distribution to mitigate noise interference in predictions. By simultaneously minimizing prediction entropy and maximizing inter-prototype distances, CRG achieves multi-modal alignment with enhanced robustness and generalization.}
    \label{fig:architechture}
\end{figure*}
\textbf{Test-Time Logit Inference and Loss Function.}

The learning diagram of CRG can be illustrated by Fig.\ref{fig:architechture}. Given the complexity of Gaussian Discriminant Analysis and the risk of instability in high-dimensional spaces, we continue to use the similarity matching method to adjust class prototypes. Using the text cache, positive cache prototypes, and negative cache prototypes, the prediction for the input sample $I$ is as follows:

\begin{equation}
P(c_k \mid I) = \frac{\exp\left((f_I^\top t_k + \mathcal{A}(f_I^\top v_k^+) + \mathcal{B}(f_I^\top v_k^-)) / \tau \right) }{\sum_{j=1}^K \exp\left((f_I^\top t_j + \mathcal{A}(f_I^\top v_j^+) + \mathcal{B}(f_I^\top v_j^-)) / \tau \right) }
\label{logits}
\end{equation}

Here, $\top$ denotes the matrix transpose, $\mathcal{A}(x) = \lambda_1 \exp(-\beta(1 - x))$ is the logits for positive sample inference, where $\lambda_1$ is the balance parameter for positive samples and $\beta$ is the sharpness ratio; $\mathcal{B}(x) = \lambda_2 \exp(\beta(1 - x)) $ represents the logits for negative sample inference, where $\lambda_2$ is the balance parameter for negative samples.

Similar to the loss function used in TPT\cite{TPT}, we optimize these three residuals to promote consistent predictions across \( N \) different augmented views of the given test image \( I \) using the unsupervised entropy minimization objective.
 \begin{equation}
    \mathcal{L}_{\text{TPT}} = H\left( \frac{1}{\rho N} \sum_{i=1}^N \mathbb{I}\left( H(p(y \mid \tilde{I}_i)) \leq \theta \right) P(y \mid \tilde{I}_i) \right),
 \label{test-time prompt}
 \end{equation}

 where \( p(y \mid \tilde{I}_i) \) is the predicted probability distribution for the augmented view \( \tilde{I}_i \), \( H(\cdot) \) is the entropy function, \( \theta \) is an entropy threshold, \( \rho  \) represents the filtered ratio, and \( \mathbb{I}(\cdot) \) is the indicator function ensuring that only augmented views with low entropy are used for training.
 
However, in traditional TPT\cite{TPT}, there are no cached features. When we work with a cache that has labeled samples and test samples without labels, this essentially becomes an unsupervised domain adaptation problem.  We believe that during test time, we should aim to separate prototypes of different categories as much as possible. At the same time, we also require that positive cache prototypes and negative cache prototypes be well separated. Based on this, we define the following loss function:

\begin{equation}
\mathcal{L}_{\text{inter}}^{\text{text}} = \sum_{m \neq n} \exp\bigl(-\gamma\|t_m - t_n\|_2^2\bigr)
\label{text_inter}
\end{equation}


\begin{equation}
\mathcal{L}_{\text{positive}}^{\text{negative}} =  \sum_{c=1}^K  \frac{\langle v_c^+, v_c^- \rangle}{\|v_c^+\| \|v_c^-\|}.
\label{positive_negtive}
\end{equation}

Here, Equation~\ref{text_inter} is used to separate the text prototype. In contrast, Equation~\ref{positive_negtive} is used to separate the positive cache prototype and the negative cache prototype, where $t_m$ is the text features of the $m$-th class, $v_c^+$ denotes the positive cache prototype, and $v_c^-$ denotes the negative cache prototype.

Overall, our final optimization objective is:
\begin{equation}
\mathcal{L}_{\text{total}} = \mathcal{L}_{\text{TPT}} + \xi_1 \mathcal{L}_{\text{inter}}^{\text{text}} 
+ \xi_2 \mathcal{L}_{\text{positive}}^{\text{negative}}.
\end{equation}

where \( \xi_1\) and \( \xi_2\) are two weight hyperparamenters. We can achieve better alignments of the three different prototype features through these approaches.

\subsection{Gaussian}

\textbf{Theoretical Justification.}
Recent studies\cite{mahalanobis1930group} have theoretically demonstrated that features follow a Gaussian distribution when the network is trained with the Softmax function. Since CLIP uses the Softmax function during training, this provides a theoretical justification for applying Gaussian Discriminant Analysis\cite{bishop2006pattern}. In previous calibration processes, learning a residual essentially corresponds to applying a simple linear transformation to the features. If the original features follow a Gaussian distribution, the calibrated features will also retain the Gaussian distribution property. Furthermore, \cite{AHTB} has successfully applied GDA in few-shot, base-to-novel, and unsupervised scenarios to adapt visual-language models.

\textbf{Gaussian Discriminant Analysis.} We first introduce the assumption of GDA, where the feature \(f_I\) under class \(k\) follow a Gaussian distribution. Specifically, the class-conditional distribution can be expressed as:

\[
p(I \mid y = k) \sim \mathcal{N}(\mu_k, \Sigma),
\]

where \(\mu_k\) is the mean vector for class \(c_k\), and \(\Sigma\) is the shared covariance matrix across all classes.

Based on this assumption, we use Bayes' theorem to compute the posterior probability \(p(y = k \mid I)\), given by:

\[
p(y = k \mid I) = \frac{p(f_I \mid y = k)p(y = k)}{\sum_{j=1}^K p(f_I \mid y = j)p(y = j)}.
\]

Substituting the Gaussian form of \(p(I \mid y = k) \sim \mathcal{N}(\mu_k, \Sigma),\) into the above equation, the posterior probability can be further written as follows. The detailed derivation process is provided in the appendix. :

\[
p(y = k \mid I) = \frac{\exp\left( \mu_k^T \Sigma^{-1} f_I - \frac{1}{2} \mu_k^T \Sigma^{-1} \mu_k + \log p_k \right)}{\sum_{j=1}^K \exp\left( \mu_j^T \Sigma^{-1} f_I - \frac{1}{2} \mu_j^T \Sigma^{-1} \mu_j + \log p_j \right)},
\]

where \(p_k = p(y = k) = \frac{1}{K}\) for \(k = 1, 2, \dots, K\), assuming a uniform prior distribution across all classes.

Through normalization of the posterior probability, the classifier can be equivalently represented as a linear classifier, where the weight \(w_k\) and bias \(b_k\) are defined as:

\[
w_k = \Sigma^{-1} \mu_k, \quad b_k = \log p_k - \frac{1}{2} \mu_k^T \Sigma^{-1} \mu_k.
\]

Finally, the decision function for the classifier is given by:

\begin{equation}
{h}_k(f_I) = w_k^T f_I + b_k.
\label{gda}
\end{equation}

In the actual inference process, we use the well-aligned positive cache features to compute the mean for each class and the overall covariance, thereby constructing a Gaussian discriminant classifier. Additionally, to prevent some category queues from being empty at initialization, we insert a pair consisting of a textual vector and a high-entropy value into the queue of each category.

Finally, when providing the final inference results, we replace $\mathcal{A}(f_I^\top v_k^+)$ with $\lambda_1 {h}_k(f_I)$, following the format of Equation \ref{logits} and \ref{gda}, we get Equation \ref{eleven}.
\begin{equation}
\label{eleven}
p(y=k \mid I) = \frac{\exp\left((f_I^\top t_k + \lambda_1 {h}_k(f_I) + \mathcal{B}(f_I^\top v_k^-)) / \tau \right) }{\sum_{j=1}^K \exp\left((f_I^\top t_j + \lambda_1 {h}_i(f_I) + \mathcal{B}(f_I^\top v_j^-)) / \tau \right) }
\end{equation}

\section{Experiment}

\subsection{Datasets.} We follow previous work\cite{TDA} to evaluate our method on two benchmarking scenarios: cross-dataset generalization and robustness to natural distribution shifts. (1) For cross-dataset generalization tasks, we conduct comprehensive assessments across 10 diverse recognition datasets, 
including FGVC Aircraft\cite{aircraft}, Caltech101\cite{caltech101}, Stanford Cars\cite{cars}, DTD\cite{Dtd}, EuroSAT\cite{helber2019eurosat}, Flowers102\cite{oxfordflower}, Food101\cite{bossard2014food}, Oxford Pets\cite{oxford_pets}, SUN397\cite{sun397}, and UCF101\cite{ucf101}. 
These datasets provide a comprehensive benchmark for evaluating the adaptability and generalization ability of methods across different datasets. (2) For evaluating robustness to natural distribution shifts, we assess the performance of our method using the 
ImageNet\cite{deng2009imagenet} dataset alongside its variant out-of-distribution datasets, including ImageNet-A\cite{hendrycks2021natural}, ImageNet-V2\cite{recht2019imagenet}. 
This evaluation measures our method's robustness in the presence of different distribution shifts. For datasets with excessively large test sets—Food101, ImageNet, SUN397, ImageNet-V2, and ImageNet-A—we randomly sample 2000 test instances for evaluation, following the experimental setting in DiffTPT\cite{Feng2023DiverseDA}.

\begin{wraptable}{r}{0.6\linewidth} 
\vspace{-10pt} 
\renewcommand\arraystretch{0.9}
\renewcommand{\tabcolsep}{1pt}
\centering
\caption{\textbf{Performance comparisons on robustness to natural distribution shifts}. 
The best results are highlighted in \textbf{bold}. 
}
\label{tab:ood-main}
\vspace{-5pt}
\scriptsize
\resizebox{1\linewidth}{!}{
  \begin{tabular}{llcccc} 
    \toprule
    {Method} & {Publication} & ImageNet & -A & -V2 & {Average} \\
    \midrule
    CLIP-ResNet-50 & ICML 2021 & 58.16 & 21.83 & 56.15 & 44.18 \\ 
    \midrule
    CoOp~\cite{CoOp} & IJCV 2022 & 63.33 & 23.06 & \textbf{56.60} & 46.66 \\
    \midrule
    TPT~\cite{TPT} & NIPS 2022 & 60.74 & 26.67 & 54.70 & 48.84 \\
    DiffTPT~\cite{Feng2023DiverseDA} & ICCV 2023 & 60.80 & \textbf{31.06} & 55.80 & 49.22 \\
    TDA~\cite{TDA} & CVPR 2024 & 61.35 & 30.29 & 55.54 & 49.06 \\ 
    DPE~\cite{DPE} & NIPS 2024 & 63.41 & 30.15 & \textbf{56.72} & 50.09 \\ 
    \rowcolor{gray!20}
    \textbf{Ours} & ICME 2025 & \textbf{65.26} & 29.69 & 56.07 & \textbf{50.34} \\ 
    \midrule
    \midrule
    CLIP-ViT-B/16 & ICML 2021 & 66.73 & 47.87 & 60.86 & 58.49 \\
    \midrule
    CoOp~\cite{CoOp} & IJCV 2022 & 71.51 & 49.71 & 64.20 & 61.81 \\
    \midrule
    TPT~\cite{TPT} & NIPS 2022 & 68.98 & 54.77 & 63.45 & 62.40 \\
    DiffTPT~\cite{Feng2023DiverseDA} & ICCV 2023 & 70.30 & 55.68 & 65.10 & 63.69 \\
    TDA~\cite{TDA} & CVPR 2024 & 69.51 & 60.11 & 64.67 & 64.76 \\ 
    DPE~\cite{DPE} & NIPS 2024 & 71.91 & 59.63 & \textbf{65.44} & 65.66 \\
    \rowcolor{gray!20}
    \textbf{Ours} & ICME 2025 & \textbf{75.01} & \textbf{63.67} & 64.66 & \textbf{67.78} \\ 
      \bottomrule
  \end{tabular}
}
\vspace{-10pt}
\end{wraptable}
\subsection{Implementation details}
We adopt ResNet-50 and ViT-B/16 as the visual encoders for CLIP. 
We use hand-crafted prompt, which are detailed in the appendix. Following the approach of TPT~\cite{TPT}, we generate 63 augmented views for each test image. 
For the learning of three residual parameters, we utilize the AdamW optimizer with a learning rate of 0.0005, completing the optimization in a single iteration. 
In default, the hyperparameters are set as follows: $\xi_1$ and $\xi_2$ are set to 1 and 10, $\lambda_1$ and $\lambda_2$ are set to 7 and 0.3,  $ \tau_t $ is set to 0.1, $\rho$ is set to 0.1, $\beta$ is set to 5.0, and the queue size $\mathcal{M}$ is 12, much larger than TDA’s, since GDA is more robust to noisy pseudo-labels and additional samples enhance class distribution modeling.
All experiments are conducted on a single NVIDIA GTX 4090 GPU with 24GB of memory.

\subsection{Comparisons with State-of-the-art}

\textbf{Robustness to Natural Distribution Shifts.}
Our approach consistently demonstrates strong generalization performance on downstream tasks with natural distribution shifts. As shown in Table \ref{tab:ood-main}, our method achieved the highest average results across datasets with significant distribution differences, each evaluated using different backbone networks. Specifically, we achieved the best performance on ImageNet, with improvements of 1.85 and 3.1 percentage points. It is noteworthy that the CoOp results were obtained using few-shot learning. Our method not only surpasses TDA and DPE but, in certain cases, also outperforms few-shot methods, showcasing the superiority of our approach. Although there are instances where our performance is not as strong as DiffTPT, we speculate this may be due to the lack of extensive data augmentation. Our method can be integrated with data augmentation techniques, but this was not the primary focus of our study.

\begin{table*}[t]
\renewcommand\arraystretch{1.3}
\renewcommand{\tabcolsep}{2pt}
  \caption{\textbf{Performance comparisons on cross-datesets generalization.} The best results are highlighted in \textbf{bold}.}
  \vspace{0pt}
  \centering
  \resizebox{1\linewidth}{!}{
    \begin{tabular}{lp{1.2cm}<{\centering}p{1.2cm}<{\centering}p{1.2cm}<{\centering}p{1.2cm}<{\centering}p{1.2cm}<{\centering}p{1.2cm}<{\centering}p{1.2cm}<{\centering}p{1.2cm}<{\centering}p{1.2cm}<{\centering}p{1.2cm}<{\centering}p{1.2cm}<{\centering}}
      \toprule
      Method & Aircraft & Caltech & Cars & DTD & EuroSAT & Flower & Food101 & Pets & SUN397 & UCF101 & Average \\
      \midrule
      CLIP-ResNet-50 & 15.66 & 85.88 & 55.70 & 40.37 & 23.69 & 61.75 & 73.97 & 83.57 & 58.80 & 58.84 & 55.82 \\
      \midrule
      CoOp~\cite{CoOp} & 15.12 & 86.53 & 55.32 & 37.29 & 26.20 & 61.55 & 75.59 & \textbf{87.00} & 58.15 & 59.05 & 56.18 \\
      \midrule
      TPT~\cite{TPT}  & 17.58 & 87.02 & 58.46 & 40.84 & 28.33 & 62.69 & 74.88 & 84.49 & 61.46 & 60.82 & 57.66 \\
      DiffTPT~\cite{Feng2023DiverseDA} & 17.60 & 86.89 & \textbf{60.71} & 40.72 & 41.04 & 63.53 & \textbf{79.21} & 83.40 & 62.72 & 62.67 & 59.85 \\
      TDA~\cite{TDA}  & 17.61 & 89.70 & 57.78 & 43.74 & 42.11 & 68.74 & 77.75 & 86.18 & 62.53 & \textbf{64.18}  & 61.03\\ 
       DPE~\cite{DPE}  & \textbf{19.80} & \textbf{90.83} & 59.26 & 50.18 & 41.67 & 67.60 & 77.83 & 85.97 & \textbf{64.23} & 61.98  & 61.93\\ 
      \rowcolor{gray!20}
      \textbf{Ours} & 18.09 & 90.12 & 57.92 & \textbf{51.89} & \textbf{46.80} & \textbf{71.09} & 75.76 & 85.91 & 63.11 & 63.58 & \textbf{62.42} \\
      \midrule
      \midrule
      CLIP-ViT-B/16 & 23.67 & 93.35 & 65.48 & 44.27 & 42.01 & 67.44 & 83.65 & 88.25 & 62.59 & 65.13 & 63.58 \\
      \midrule
      CoOp~\cite{CoOp} & 18.47 & 93.70 & 64.51 & 41.92 & 46.39 & 68.71 & 85.30 & 89.14 & 64.15 & 66.55 & 63.88 \\
      \midrule
      TPT~\cite{TPT}  & 24.78 & 94.16 & 66.87 & 47.75 & 42.44 & 68.98 & 84.67 & 87.79 & 65.50 & 68.04 & 65.10 \\
      DiffTPT~\cite{Feng2023DiverseDA} & 25.60 & 92.49 & 67.01 & 47.00 & 43.13 & 70.10 & \textbf{87.23} & 88.22 & 65.74 & 62.67 &65.47 \\
      TDA~\cite{TDA} & 23.91 & 94.24 & 67.28 
      & 47.40 & 58.00 & 71.42 & 86.14 & 88.63 & 67.62 & \textbf{70.66} & 67.53 \\ 
      DPE~\cite{DPE} & \textbf{28.95} & \textbf{94.81} & \textbf{67.31} 
      & 54.20 & 55.79 & 75.07 & 86.17 & 91.14 & \textbf{70.07} & 70.44 & 69.40 \\ 
      \rowcolor{gray!20}
      \textbf{Ours} & 26.58 & 93.57 & 66.89 & \textbf{56.38} & \textbf{59.81} & \textbf{75.94} & 85.95 & \textbf{91.20} & 68.36 & 70.31 & \textbf{69.50} \\
      \bottomrule
    \end{tabular}
  } 
  \label{tab:fine-grained}
  \vspace{-10pt}
\end{table*}

\textbf{Cross-Datasets Generalization.}
In Table \ref{tab:fine-grained}, we further evaluate the generalization capability of our proposed method against other state-of-the-art methods across 10 fine-grained recognition datasets. Due to the substantial distribution differences among these datasets, performance can vary considerably. Despite this, our method achieves average improvements of 0.49 over the current best-performing methods on the CLIP-ResNet-50 backbone, surpassing them on 3 out of the 10 datasets. On the CLIP-ViT backbone, our method’s average accuracy is 0.1 higher than DPE, and is optimal on 4 of the 10 datasets. Notably, we significantly improved EuroSAT due to its more compact category distributions, yielding more representative prototypes. However, our performance on Food101 is slightly lower, likely due to the high inter-class similarity that poses challenges for GDA. Overall, these findings confirm our method’s robustness and adaptability when transferring to new domains at test time, making it crucial for real-world applications.

\subsection{Discussion}
\begin{wraptable}{r}{0.5\linewidth} 
\vspace{-10pt} 
\renewcommand\arraystretch{1.2}
\renewcommand{\tabcolsep}{2pt}
\caption{\looseness=-1 \textbf{Ablation studies for different variants of our method}.}
\label{table:ablation}
\centering
\resizebox{\linewidth}{!}{%
  \begin{tabular}{cccccc|ccc}
    \toprule
    ${T}_{\text{cache}}$ & $V_{\text{cache}}^+$ & $V_{\text{cache}}^-$ & $GDA$ & $\mathcal{L}_{\text{inter}}^{\text{text}}$ & $\mathcal{L}_{\text{positive}}^{\text{negative}}$ & Flowers  & ImageNet\\
    \midrule
    \ding{51} & \ding{51} &  \ding{55}   & \ding{55} &  \ding{55}   & \ding{55} &  73.35 &  73.27  \\
    \ding{51} & \ding{51} &  \ding{55}   & \ding{55} &  \ding{51}   & \ding{55} &  73.57 &  73.56  \\
    \ding{51} & \ding{51} &  \ding{51}   &  \ding{55}  &  \ding{55}   & \ding{55} & 73.35  &  73.37 \\
    \ding{51} & \ding{51} &  \ding{51}   &  \ding{55}  &  \ding{55}   & \ding{51} & 73.57  & 73.66  \\
    \ding{51} & \ding{51} &  \ding{55}   &  \ding{51} &  \ding{55}   & \ding{55} & 75.19   &  74.27  \\ 
    \rowcolor{gray!20}
    \ding{51} & \ding{51} &  \ding{51}   &  \ding{51}  &  \ding{51}   & \ding{51} & \textbf{75.94}  &  \textbf{75.01} \\
    \bottomrule
  \end{tabular}%
}
\vspace{-10pt} 
\end{wraptable}
\textbf{Ablation Analysis.} 
To further analyze the effectiveness of our method, we conducted an ablation study to examine the impact of different components in Table \ref{table:ablation}. Simply adding the negative cache and introducing learnable residual parameters may lead to ineffective parameter learning if there is no robust representation or adequate constraint guidance, which in turn can result in misalignment between different modalities. As observed, the performance change in this case is nearly zero. By introducing GDA and imposing constraints, we not only achieve more accurate inference but also facilitate effective prototype alignment. The ablation results demonstrate that these modules work together to fully leverage the trustworthy cache mechanism, ultimately improving the overall accuracy of the model.

\textbf{Why Is GDA Robust?} From the perspective of Bayesian decision theory\cite{hart2000pattern}, the GDA classifier is an approximate implementation of the minimum error rate classifier based on Bayesian theory\cite{berger1994overview}. When class distributions can be described by Gaussian functions, Bayesian decision theory indicates that the theoretical optimal classification boundary can be achieved using the means and covariances. Noise in labels introduces errors when estimating the class-conditional distributions; however, since GDA employs a global (distribution-level) modeling approach, it approximates the true distribution as a whole. This allows GDA to remain robust and approximate the optimal decision boundary even when some labels are incorrect\cite{berger1994overview}. 
\section{Conclusion}
In summary, we propose an innovative test-time adaptation framework for vision-language models by integrating Cache, Residual, and Gaussian. Through residual learning for prototype alignment, Gaussian Discriminant Analysis for category modeling, and a Negative Cache, our method excels under cross-domain and natural distribution shifts, offering a fresh perspective on test-time adaptation.

\bibliography{iclr2025_conference}
\bibliographystyle{iclr2025_conference}

\appendix
\section{Appendix}

In appendix, we provide additional details and experimental results to enhance understanding and insights into our method.
This supplementary document is organized as follows:
\begin{itemize}
     \item[\textbullet] \textbf{Detailed Dataset Information:}  
    Comprehensive details about the datasets used in our experiments, including their key characteristics and distributions, are provided.

    \item[\textbullet] \textbf{Preliminaries and Methodological Differences:}  
    A discussion of a foundational concept is included, along with an explanation of the key distinctions between our approach and TDA\cite{TDA},SimNL\cite{simNL},GDA\cite{AHTB},DPE\cite{DPE}.

    \item[\textbullet] \textbf{Text Templates for Each Dataset:}  
    The text templates used in our experiments for each dataset are listed for reproducibility.

    \item[\textbullet] \textbf{Derivation of Gaussian Discriminant Analysis (GDA):}  
    The mathematical derivation of GDA is detailed.

    \item[\textbullet] \textbf{Ablation Study on Hyperparameter $\mathcal{M}$:}  
    An analysis of the hyperparameter $M$ is presented, showing its impact on performance through ablation experiments.

    \item[\textbullet] \textbf{Analysis of $\lambda_1$ and $\lambda_2$:}  
    Insights into the tuning strategies and effects of the hyperparameters $\lambda_1$ and $\lambda_2$ are discussed.

    \item[\textbullet] \textbf{Motivation of $\mathcal{L}_{\text{inter}}^{\text{text}}$ and $\mathcal{L}_{\text{positive}}^{\text{negative}}$:}  
    The motivations behind the losses $\mathcal{L}_{\text{inter}}^{\text{text}}$ and $\mathcal{L}_{\text{positive}}^{\text{negative}}$ are analyzed, demonstrating how they help reduce overconfidence and enhance the model's ability to distinguish positive and negative prototypes.
    


\end{itemize}

\subsection{Detailed Dataset Information}

In Table\ref{tab:dataset}, we provide comprehensive statistics for each dataset utilized in our experiments, detailing the number of classes, the sizes of the training, validation, and test sets, as well as their associated original tasks. These datasets have emerged as key benchmarks for evaluating the test-time adaptation of vision-language models\cite{TPT,TPS,TDA,Feng2023DiverseDA}.

\begin{table}[ht]
    \caption{\looseness=-1 Detailed statistics of datasets used in experiments. Note that the last 2 ImageNet variant datasets are designed for evaluation and only contain the test sets. \textbf{Datasets marked with an asterisk$^*$ indicate that 2,000 samples were randomly selected for testing during the evaluation process.}}
    \large
    \label{tab:dataset}
    \centering
    \resizebox{\textwidth}{!}{
    \setlength{\tabcolsep}{2mm}{
    \begin{tabular}{lccccc}
    \toprule
Dataset                  & Classes  & Training & Validation   & Testing & Task \\ \midrule
Caltech101~\cite{caltech101} & 100 & 4,128 & 1,649 & 2,465& Object recognition \\
DTD~\cite{Dtd}& 47 & 2,820 & 1,128& 1,692 &  Texture recognition\\ 
EuroSAT~\cite{helber2019eurosat}& 10 & 13,500 & 5,400& 8,100 & Satellite image recognition \\ 
FGVCAircraft~\cite{aircraft} & 100 & 3,334 & 3,333& 3,333 & Fine-grained aircraft recognition\\
Flowers102~\cite{oxfordflower} & 102 & 4,093 & 1,633& 2,463 & Fine-grained flowers recognition \\ 
Food101$^*$~\cite{bossard2014food} & 101 & 50,500& 20,200& 30,300 & Fine-grained food recognition  \\ 
ImageNet$^*$~\cite{deng2009imagenet} & 1,000 & 1.28M & -& 50,000 & Object recognition \\ 
OxfordPets~\cite{oxford_pets} & 37  & 2,944 & 736& 3,669 & Fine-grained pets recognition \\ 
StanfordCars~\cite{cars} & 196 & 6,509 & 1,635& 8,041 & Fine-grained car recognition \\
SUN397$^*$~\cite{sun397}& 397& 15,880 & 3,970& 19,850 & Scene recognition\\ 
UCF101~\cite{ucf101}& 101 & 7,639 & 1,898& 3,783 & Action recognition\\
\midrule
ImageNet-V2$^*$~\cite{recht2019imagenet} & 1,000 & - & -& 10,000 & Robustness of collocation  \\
ImageNet-A$^*$~\cite{hendrycks2021natural}& 200 & - & -&7,500 &Robustness of adversarial attack\\
    \bottomrule
    \end{tabular}
    }
    }
\end{table}

\subsection{Preliminaries and Methodological Differences}

\textbf{Test-Time Prompt Tuning (TPT).}  
TPT\cite{TPT} is an augmentation-based adaptation method designed to enhance the generalization of pre-trained models during testing. For each test sample \( X_{\text{test}} \), TPT generates \( n \) augmented views \( \tilde{X}_i = A_i(X_{\text{test}}) \) using augmentation functions \( A \), and adapts the model with learned prompts for these views. The objective is to minimize prediction uncertainty by reducing marginal entropy across augmented views while filtering out noisy augmentations through confidence selection. The loss function is defined as:

\begin{equation}
    \mathcal{L}_{\text{TPT}} = H\left( \frac{1}{\rho N} \sum_{i=1}^N \mathbb{I}\left( H(P(y \mid E(\tilde{X}_i, \theta))) < \tau \right) P(y \mid E(\tilde{X}_i, \theta)) \right),
\label{test-time prompt}
\end{equation}

where \( P(y \mid E(\tilde{X}_i, \theta)) \) is the predicted probability distribution for the augmented view \( \tilde{X}_i \), \( E(\cdot, \theta) \) is the model with prompt tuning, \( H(\cdot) \) is the entropy function, \( \tau \) is an entropy threshold, and \( \mathbb{I}(\cdot) \) is the indicator function ensuring that only augmented views with low entropy are used for training.

It is important to note that TPT provides a new paradigm for optimization, where the specific parameters to be tuned are not fixed but can be adapted flexibly as long as the loss function is followed. 

{Our method still follows the TPT paradigm. However, unlike the TPT approach, we learn multimodal residuals, which eliminates the need for gradients to pass through the encoder, significantly improving training speed.}

\begin{figure}[!ht]
    \centering
    \begin{subfigure}{0.4\textwidth} 
        \centering
        \includegraphics[width=\linewidth]{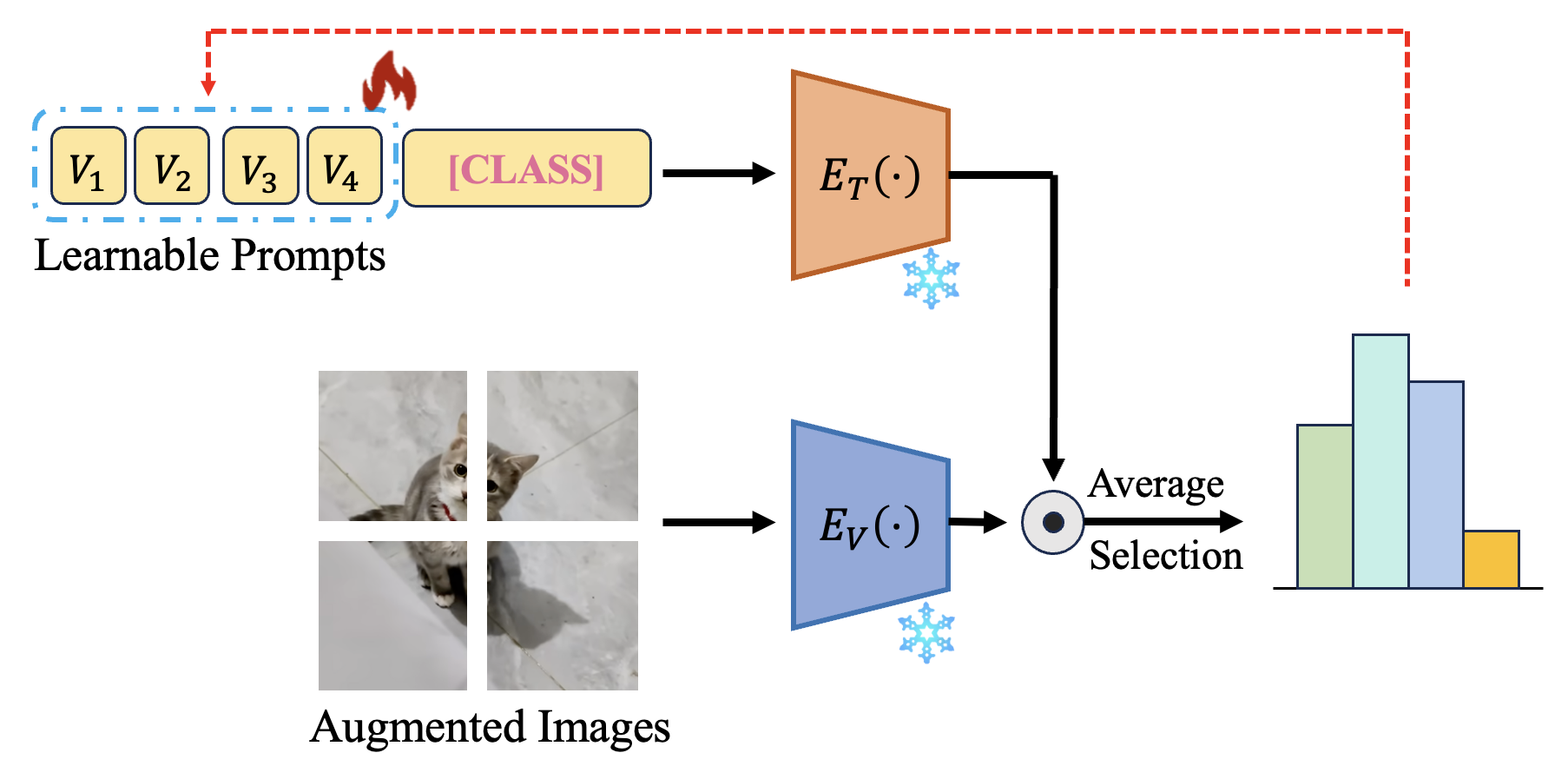}
        \caption{Test-Time Prompt Tuning}
        \label{fig:image1}
    \end{subfigure}\hfill
    \begin{subfigure}{0.53\textwidth} 
        \centering
        \includegraphics[width=\linewidth]{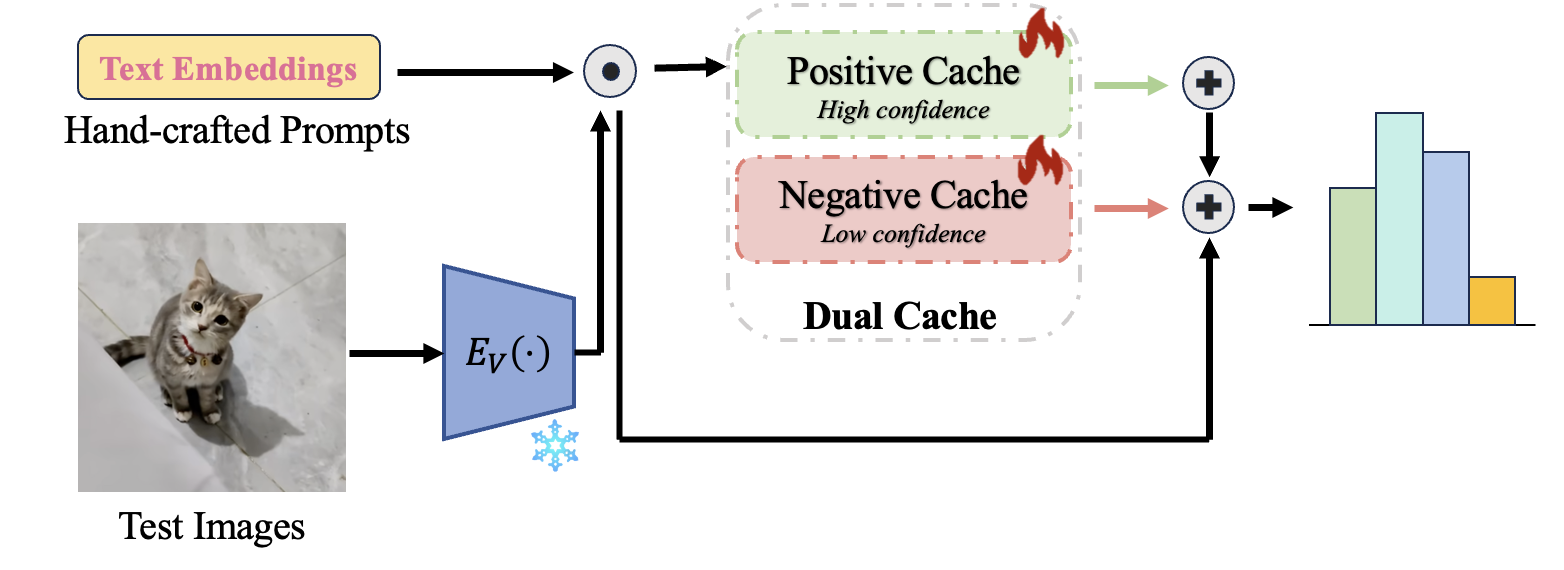}
        \caption{Training-free Dynamic Adaption}
        \label{fig:image2}
    \end{subfigure}
    
    \vspace{5mm}
    
    \begin{subfigure}{0.7\textwidth} 
        \centering
        \includegraphics[width=\linewidth]{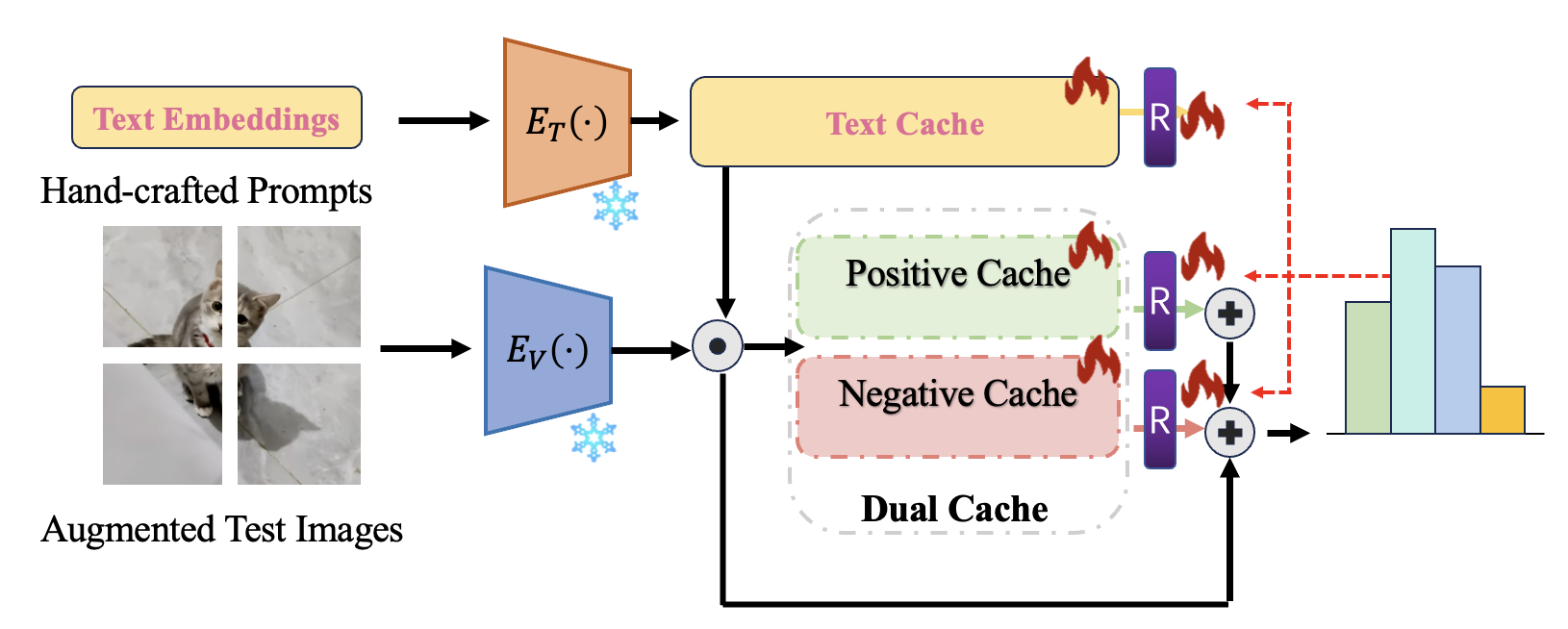}
        \caption{CRG}
        \label{fig:image3}
    \end{subfigure}

    \caption{Two classic methods of Test-Time Adaptation (top) and our cache-based approach (bottom).}
    \label{fig:difference}
\end{figure}

\textbf{Differences from TDA\cite{TDA}}  Although our approach shares many visual similarities with TDA\cite{TDA} methods, it is fundamentally different in methodology. Our Negative Cache Prototypes are entirely derived from the Positive Cache and are constructed following the principles outlined in the main text. This means we focus exclusively on high-confidence images. Building on this foundation, we further learn a bi-modal residual to better handle each test sample. Additionally, we incorporate Gaussian Discriminant Analysis during inference, which effectively mitigates the impact of noisy labels. 	The differences between our method and those of TDA and TPT are clearly illustrated in \ref{fig:difference}

\textbf{Differences from SimNL\cite{simNL}} SimNL proposes a simple yet effective negative learning method for few-shot scenarios by constructing a negative cache to learn complementary features, thereby adapting VLMs to downstream tasks. Specifically, SimNL constructs negative features on both the text and visual modalities and tunes the model by learning multimodal residuals. 

Our work is inspired by the high-level idea of SimNL; however, there are significant differences in the implementation details and underlying motivation. First, the methodological paradigm is different: SimNL belongs to few-shot methods, whereas our approach focuses on test-time adaptation. We achieve unsupervised optimization by adjusting the residual between positive and negative prototypes and by incorporating an additional loss during testing to constrain their similarity, thereby forcing the two prototypes to diverge. Second, the strategy for constructing negative samples differs: we only construct negative samples on the visual modality, while SimNL constructs negative features on both text and visual modalities. Finally, regarding the construction of negative prototypes, SimNL obtains them by randomly sampling images from the \(C-1\) classes other than the target class and averaging them, repeating this process \(K\) times to form a negative cache of size \(CK \times D\) (the same shape as the positive sample cache), and then fusing the residuals via broadcasting for learning, whereas our method directly computes the negative prototype from the positive prototype, resulting in a negative prototype of size \(C \times D\). Moreover, our motivation for introducing negative learning is to suppress label noise, which further distinguishes our approach from that of SimNL.

\textbf{Differences from GDA\cite{AHTB}} The work in \cite{AHTB} is the first to introduce Gaussian Discriminant Analysis (GDA) into the adaptation of VLMs. We acknowledge that there is a degree of similarity between their approach and ours, as both employ GDA. However, there are several important differences in our respective settings, motivations, and technical details. First, \cite{AHTB} focuses on few-shot scenarios, extending to base-to-new class generalization and unsupervised learning, whereas our method is designed for test-time adaptation using a cache-based strategy. Second, while \cite{AHTB} primarily adopts GDA as a training-free solution to minimize additional parameters and computational overhead, we leverage GDA from a distributional perspective to enhance robustness to noisy labels—a notable advantage over KNN-like approaches\cite{TipAdapter}. Finally, when computing the mean and covariance for each class, we incorporate a learned residual from test-time adaptation: for each image in the cache, we add the learned residual to its representation before deriving the class mean and covariance. This residual-based adjustment further distinguishes our method from that of \cite{AHTB}.

\textbf{Difference form DPE\cite{DPE}}
Residual learning has demonstrated great potential in the adaptation of VLMs. Taskres\cite{TaskRes}, in few-shot scenarios, achieves rapid adaptation to downstream tasks by learning only a single layer of prior-agnostic parameters on the text modality. TPS\cite{TPS} extends this strategy further into test-time adaptation, adhering to the TPT\cite{TPT} paradigm by continuously updating text prototypes during testing. On the other hand, DPE\cite{DPE} integrates cache-based approaches with residual learning, introducing for the first time the concept of Dual Prototype Evolution in test-time adaptation. Specifically, DPE learns residuals for both textual and visual prototypes simultaneously, dynamically updating prototypes across the two modalities throughout testing. This makes test-time adaptation of vision-language models simultaneously accumulative and multimodal for the first time.

Our work is inspired by the overall framework of DPE and similarly adopts the dynamic updating mechanism for text and visual prototypes. However, our method further introduces negative prototype learning. Specifically, during test-time adaptation, we not only aim to enlarge the distinction among different textual prototypes but also explicitly reduce the similarity between positive and negative prototypes. After residual learning, we integrate the learned residuals into cached features and utilize Gaussian Discriminant Analysis for inference, thus achieving a more robust and accurate test-time adaptation performance.

\subsection{Text Templates for Each Dataset:}
\label{sec:prompt}
In Table~\ref{tab:prompt}, we detail the specific hand-crafted prompts utilized for each dataset. 

\begin{table}[t]
\centering
    \caption{\looseness=-1 \textbf{Textual prompts used in experiments}. In addition to these prompts, we also employ CuPL~\cite{Pratt2022WhatDA} prompts to further enhance performance.}
    \large
    \label{tab:prompt}
    \resizebox{0.9\textwidth}{!}{
    \setlength{\tabcolsep}{8mm}{
    \begin{tabular}{lc}
    \toprule
Dataset                  & Prompts   \\ \midrule
& ``itap of a \{\texttt{CLASS}\}.'' \\ 
& ``a bad photo of the \{\texttt{CLASS}\}.'' \\ 
ImageNet~\cite{deng2009imagenet}& ``a origami \{\texttt{CLASS}\}.'' \\ 
ImageNet-V2~\cite{recht2019imagenet}& ``a photo of the large \{\texttt{CLASS}\}.'' \\ 
ImageNet-A~\cite{hendrycks2021natural}& ``a \{\texttt{CLASS}\} in a video game.'' \\ 
& ``art of the \{\texttt{CLASS}\}.'' \\ 
& ``a photo of the small \{\texttt{CLASS}\}.'' \\ \midrule
Caltech101~\cite{caltech101} & ``a photo of a \{\texttt{CLASS}\}.'' \\
DTD~\cite{Dtd}& ``\{\texttt{CLASS}\} texture.'' \\ 
EuroSAT~\cite{helber2019eurosat}& ``a centered satellite photo of \{\texttt{CLASS}\}.'' \\ 
FGVCAircraft~\cite{aircraft} & ``a photo of a \{\texttt{CLASS}\}, a type of aircraft.'' \\
Flowers102~\cite{oxfordflower} & ``a photo of a \{\texttt{CLASS}\}, a type of flower.'' \\ 
Food101~\cite{bossard2014food} & ``a photo of \{\texttt{CLASS}\}, a type of food.'' \\ 
OxfordPets~\cite{oxford_pets} & ``a photo of a \{\texttt{CLASS}\}, a type of pet.''  \\ 
StanfordCars~\cite{cars} & ``a photo of a \{\texttt{CLASS}\}.'' \\
SUN397~\cite{sun397}& ``a photo of a \{\texttt{CLASS}\}.''\\ 
UCF101~\cite{ucf101}& ``a photo of a person doing \{\texttt{CLASS}\}.'' \\
    \bottomrule
    \end{tabular}
    }
    }
\end{table}

\subsection{Derivation of Gaussian Discriminant Analysis (GDA)}
In the main text, we use Bayes’ theorem to compute the posterior probability. By substituting the Gaussian form \(p(x \mid y = i) \sim \mathcal{N}(\mu_i, \Sigma)\) into the posterior probability formula, we arrive at the following expression for \(p(y = i \mid x)\), which can be represented by a linear classifier. Below, we provide a more detailed illustration of how this posterior probability formula is derived.

Given \(p(x \mid y = i) \sim \mathcal{N}(\mu_i, \Sigma)\), the likelihood of class \(i\) is
\begin{equation}
\label{eq:multi_gaussian}
    p(x \mid y = i) = \frac{1}{(2\pi)^{\frac{d}{2}} \lvert \Sigma \rvert^{\frac{1}{2}}}
    \exp \Bigl(-\frac{(x - \mu_i)^T \Sigma^{-1} (x - \mu_i)}{2}\Bigr),
\end{equation}
where \(d\) is the dimension of the feature vector.

Using Bayes’ theorem, the posterior probability \(p(y = i \mid x)\) can be written as
\begin{equation}
\begin{aligned}
    p(y = i \mid x) 
    &= \frac{p(x \mid y = i)\,p(y = i)}{\sum_{j=1}^{K} p(x \mid y = j)\,p(y = j)} \quad \text{(Bayesian formula)} \\
    &= \frac{
        \frac{1}{(2\pi)^{\frac{d}{2}} \lvert \Sigma \rvert^{\frac{1}{2}}}
        \exp\Bigl(-\frac{(x - \mu_i)^T \Sigma^{-1} (x - \mu_i)}{2}\Bigr)
        p(y = i)
    }{
        \sum_{j=1}^{K} \frac{1}{(2\pi)^{\frac{d}{2}} \lvert \Sigma \rvert^{\frac{1}{2}}}
        \exp\Bigl(-\frac{(x - \mu_j)^T \Sigma^{-1} (x - \mu_j)}{2}\Bigr)
        p(y = j)
    } \quad \text{(Using Equation~\ref{eq:multi_gaussian})} \\
    &= \frac{
        \exp\Bigl(\mu_i^T \Sigma^{-1} x - \tfrac{1}{2}\mu_i^T \Sigma^{-1} \mu_i\Bigr) \, p(y = i)
    }{
        \sum_{j=1}^{K} 
        \exp\Bigl(\mu_j^T \Sigma^{-1} x - \tfrac{1}{2}\mu_j^T \Sigma^{-1} \mu_j\Bigr)\, p(y = j)
    } \quad \text{(denoted } p(y = i) = p_i\text{)} \\
    &= \frac{
        \exp\bigl(\mu_i^T \Sigma^{-1} x - \tfrac{1}{2}\mu_i^T \Sigma^{-1}\mu_i + \log p_i\bigr)
    }{
        \sum_{j=1}^K \exp\bigl(\mu_j^T \Sigma^{-1} x - \tfrac{1}{2}\mu_j^T \Sigma^{-1}\mu_j + \log p_j\bigr)
    },
\end{aligned}
\end{equation}
where we have omitted constant factors that cancel out in the ratio and combined \(\log p(y = i)\) with the exponential term, making the resulting expression directly comparable to a linear classifier.

\subsection{Ablation Study on Hyperparameter $\mathcal{M}$}
In our study, the hyperparameter $\mathcal{M}$ plays a crucial role. By default, we set $\mathcal{M}$ to 12, which is significantly larger than the value of 3 commonly used in TDA. This difference arises because Gaussian Discriminant Analysis (GDA) requires calculating the class distribution, particularly the covariance matrix and its inverse. These computations are prone to numerical instability, especially at the beginning of testing, where the classification probabilities calculated by the GDA classifier often result in NaN values.

The robustness of GDA also enables us to use a larger $\mathcal{M}$, as it effectively mitigates the impact of noisy labels. To further investigate, we conducted an analysis on the size of $\mathcal{M}$, and the results are presented below. These experiments were conducted using the RN50 backbone on ImageNet-V2.

Our conclusion is that the size of $\mathcal{M}$ should either be moderately small to maintain the purity of the cache or relatively large to enable GDA to better model the class distributions.

\begin{table}[ht]
    \centering
    \caption{Ablation study on the size of $\mathcal{M}$ and its impact on performance on ImageNet-V2.}

    \label{tab:ablation_M}
    \begin{tabular}{cccccc}
        \toprule
        \footnotesize
        \textbf{Size of $\mathcal{M}$} & 8 & 10 & 12 & 14 \\ 
        \midrule
        \footnotesize
        \textbf{Performance (\%)} & 55.6 & 55.03 & \textbf{56.07} & 55.8 \\ 
        \bottomrule
    \end{tabular}
\end{table}

\subsection{Analysis of $\lambda_1$ and $\lambda_2$}
We perform an ablation study on $\lambda_1$ and $\lambda_2$. These two parameters are responsible for adjusting the computed positive logits and negative logits, respectively. 

As shown in Table~\ref{tab:ablation_lambdas}, on the SUN397 dataset, increasing $\lambda_1$ from 5 to 7 slightly improves performance, reaching a peak at 63.11\%, while further increasing it to 10 results in a slight performance drop. This indicates that $\lambda_1$ requires careful tuning to achieve optimal results. On the other hand, on the Flowers102 dataset, performance increases steadily as $\lambda_2$ grows from 0.2 to 0.3, peaking at 71.09\%, before slightly dropping at $\lambda_2 = 0.4$. This demonstrates that both $\lambda_1$ and $\lambda_2$ significantly impact performance. These experiments are conducted using the RN50 backbone.

\begin{table}[ht]
    \centering
    \caption{Ablation study on $\lambda_1$ and $\lambda_2$.}
    \label{tab:ablation_lambdas}
    \begin{tabular}{cc}
        \begin{minipage}[t]{0.45\textwidth}
            \centering
            \caption*{(a) $\lambda_1$ Analysis on SUN397}
            \begin{tabular}{cccc}
                \toprule
                \footnotesize
                \textbf{$\lambda_1$} & 5 & 7 & 10 \\ 
                \midrule
                \footnotesize
                \textbf{Performance (\%)} & 62.96 &  \textbf{63.11} & 62.71 \\ 
                \bottomrule
            \end{tabular}
        \end{minipage}
        &
        \begin{minipage}[t]{0.45\textwidth}
            \centering
            \caption*{(b) $\lambda_2$ Analysis on Flowers102}
            \begin{tabular}{cccc}
                \toprule
                \footnotesize
                \textbf{$\lambda_2$} & 0.2 & 0.3 & 0.4 \\ 
                \midrule
                \footnotesize
                \textbf{Performance (\%)} & 70.84 & \textbf{71.09} & 70.96 \\ 
                \bottomrule
            \end{tabular}
        \end{minipage}
    \end{tabular}
\end{table}

\subsection{Motivation of $\mathcal{L}_{\text{inter}}^{\text{text}}$ and $\mathcal{L}_{\text{positive}}^{\text{negative}}$}

In this section, we discuss our motivation for designing the losses $\mathcal{L}_{\text{inter}}^{\text{text}}$ and $\mathcal{L}_{\text{positive}}^{\text{negative}}$. Specifically, $\mathcal{L}_{\text{inter}}^{\text{text}}$ enforces separations between text prototypes on a hypersphere by minimizing a Gaussian potential kernel $G$. This approach helps reduce the model’s overconfidence during test-time adaptation, thereby lowering the expected calibration error (ECE). Although a prior study \cite{yoon2024c} suggests that separating text prototypes may degrade performance, however , it brings a slight improvement in performance within our framework. Because our framework differs in two key aspects. First, we anchor the prototypes on a hypersphere by leveraging the Gaussian potential kernel, which provides a smoother separation and better preserves local geometry. Second, our overarching motivation is to control overconfidence stemming from noisy labels, and the corresponding entropy-based Priority Queue cache mechanism effectively filters out high-uncertainty samples. Consequently, once the model’s uncertainty is brought under control, the reduction in prediction entropy not only improves calibration but also contributes to slight yet consistent accuracy gains.

Meanwhile, $\mathcal{L}_{\text{positive}}^{\text{negative}}$ explicitly decreases the similarity between positive and negative cached prototypes by accentuating the distinction between “what something is” and “what something is not.” Through this contrastive mechanism, the model gains a clearer understanding of how to discriminate and recognize target concepts, thus further enhancing test-time performance under distribution shifts.

\end{document}